\documentclass[runningheads]{llncs}

\usepackage{accv}

\usepackage{accvabbrv}

\usepackage{graphicx}
\usepackage{booktabs}

\usepackage[accsupp]{axessibility}  

\usepackage[breaklinks,colorlinks,citecolor=accvblue]{hyperref}

\usepackage{orcidlink}

\usepackage{amsmath}
\usepackage{amssymb}
\usepackage{xcolor}         
\usepackage{graphicx}
\usepackage{tikz}
\usepackage{color, colortbl}
\usepackage{dsfont}
\usepackage{multirow} 
\usepackage[referable]{threeparttablex}
\usepackage{overpic}
\usepackage{wrapfig}
\usepackage{subcaption}
\usepackage{xspace}
\usepackage{enumitem}

\newcommand{\ours}{SubViT\xspace}

\definecolor{green}{HTML}{0aa344}
\definecolor{red}{HTML}{c93756}
\definecolor{darkgreen}{HTML}{068C52}
\definecolor{fullgreen}{rgb}{0.502, 0.788, 0.643}
\definecolor{fullred}{rgb}{0.800, 0.447, 0.541}
\definecolor{lightgreen}{RGB}{225, 239, 217}
\definecolor{lightblue}{RGB}{203, 220, 235}
\definecolor{fullgray}{RGB}{219, 223, 234}
\definecolor{fullpurple}{RGB}{205, 193, 255}
\definecolor{darkred}{RGB}{204, 114, 138}
\definecolor{darkpurple}{RGB}{171, 151, 255}
\definecolor{darkgray}{RGB}{114, 114, 114}

\definecolor{teal}{HTML}{14B8A6}
\definecolor{sky}{HTML}{38BDF8}
\definecolor{indigo}{HTML}{6366F1}
\definecolor{navy}{HTML}{1E3A8A}

\definecolor{amber}{HTML}{F59E0B}
\definecolor{coral}{HTML}{FF6B6B}
\definecolor{peach}{HTML}{FFB4A2}

\definecolor{sage}{HTML}{A8C3A1}
\definecolor{dustyblue}{HTML}{9BB4C7}
\definecolor{mauve}{HTML}{BFA6C9}
\definecolor{clay}{HTML}{C9B29B}

\makeatletter
\DeclareRobustCommand\onedot{\futurelet\@let@token\@onedot}
\def\@onedot{\ifx\@let@token.\else.\null\fi\xspace}
\def\eg{\emph{e.g}\onedot} 
\def\ie{\emph{i.e}\onedot}

\makeatother

\providecommand{\citep}{\cite}
\providecommand{\citet}{\cite}

\begin{document}

\title{Subtoken Vision Transformer for Fine-grained Recognition}



\author{
Jie Zhu\inst{1} \and
Ivy Zhang\inst{2} \and
Minchul Kim\inst{1} \and
Xiaoming Liu\inst{1,3}
}


\institute{
\textsuperscript{1} University of North Carolina at Chapel Hill 
\quad
\textsuperscript{2} Cranbrook Kingswood School\\
\textsuperscript{3} Michigan State University\\
\email{jiezhu@cs.unc.edu \quad izhang27@cranbrook.edu \quad kimminc2@msu.edu \quad liuxm@cs.unc.edu }
}
\maketitle

\begin{abstract}
We present Subtoken Vision Transformer (\ours), a selective image tokenization method for fine-grained visual recognition. Standard Vision Transformers compress each fixed-size patch into a single token, although fine-grained distinctions often depend on localized variations within only a few patches. \ours addresses this mismatch by representing discriminative patches with multiple subtokens while retaining the original token sequence for global context, thereby allocating additional capacity where it is most needed. Since attention heads encode complementary semantics and extracting attention maps at inference requires an extra backbone forward, we adopt a two-stage training strategy. Stage 1 fine-tunes the ViT using subdivision regions sampled from random attention heads, exposing the model to diverse subdivision patterns. Stage 2 identifies informative attention maps through feature-degradation distances and distills them into a lightweight single-map router, which directly predicts deterministic token-importance scores without a separate attention forward. We evaluate \ours on Generalized Category Discovery (GCD), a challenging task requiring both fine-grained discrimination and generalization to unlabeled novel categories. Across CUB, FGVC-Aircraft, and Stanford-Cars, \ours improves the average novel-category accuracy of DINOv2 from $81.3\%$ to $84.7\%$, with only $0.50$ ms additional latency and $3.4\%$ more FLOPs, while reducing latency by $73.8\%$ relative to Retina Patch. Code: \href{https://github.com/jiezhu23/SubViT_ACCV26}{SubViT}.
\end{abstract}

\section{Introduction} \label{sec:intro}

\begin{figure}[t]
    \centering
    \includegraphics[width=0.9\linewidth]{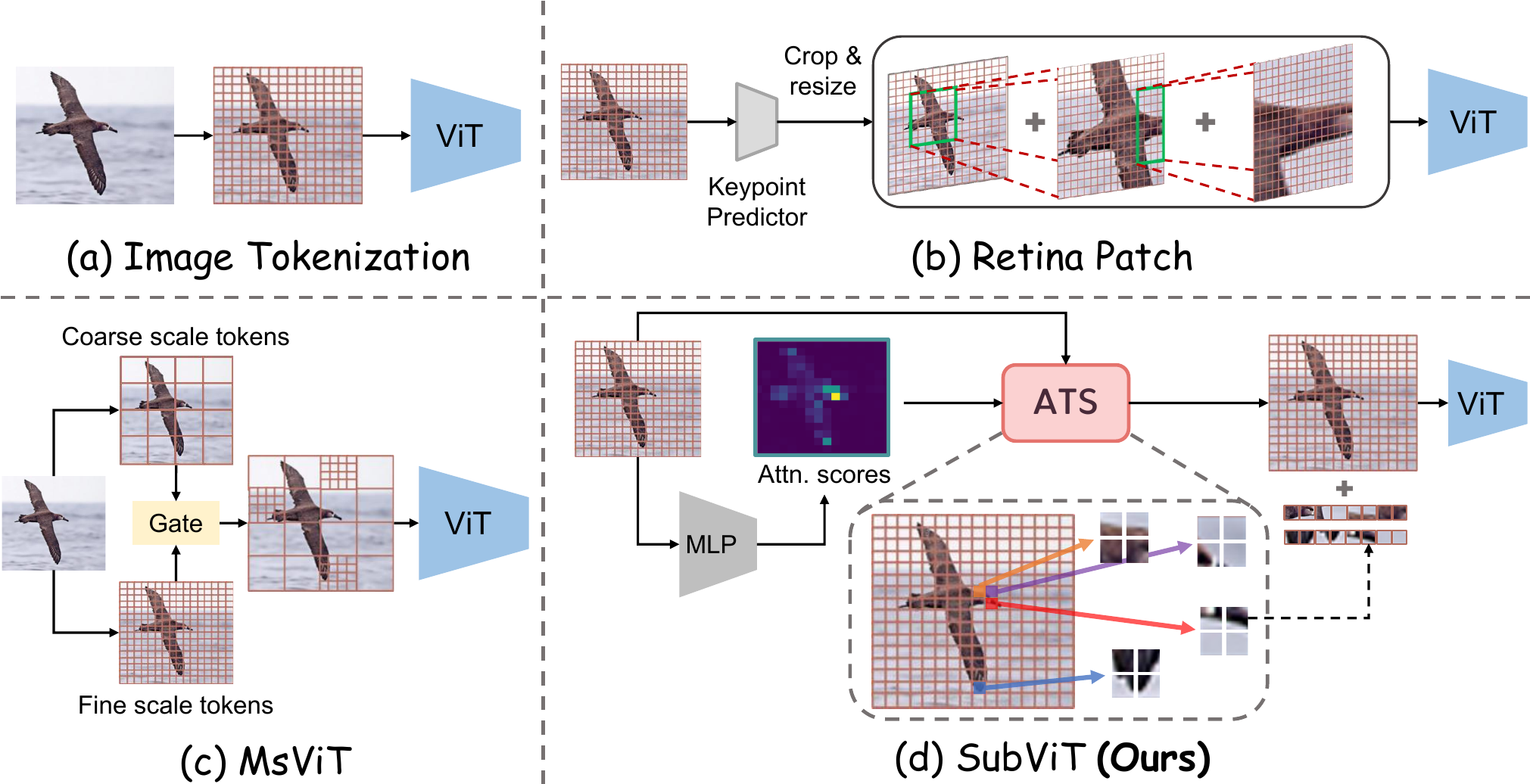}
    \caption{Motivation of \ours. Instead of densely tokenizing multiple scales, \ours identifies discriminative patches using attention supervision in training and a learned router at inference. Attention-based Token Subdivision (ATS) allocates extra subtokens only to selected patches while retaining the original tokens for global context.}
    \label{fig:intro}
\vspace{-0.3em}
\end{figure}

Vision Transformers (ViTs)~\citep{dosovitskiy2020image} represent an image as a sequence of patch tokens and use self-attention to model their global relationships. Standard ViTs uniformly partition the image into fixed-size patches and project every patch into a single token. This design is effective for capturing global structure, but it imposes the same representational granularity on all regions: spatial variations within each patch are compressed into one embedding regardless of their semantic importance. Such uniform allocation is poorly matched to fine-grained recognition~\cite{wah_branson_welinder_perona_belongie_2011, maji2013fine, krause20133d, nilsback2008automated, parkhi2012cats}, where the distinction between visually similar subcategories may depend on a small texture, contour, or object part. Uniformly refining the entire image could preserve more local structure, but would substantially increase the token sequence and computation. The central challenge is therefore to increase token-level representational capacity only where fine-grained evidence is likely to occur, while retaining the global context provided by the original token sequence.

Recent work has explored spatially adaptive tokenization. Retina Patch in SapiensID~\citep{kim2025sapiensid} uses a keypoint predictor to construct region-level crops and concatenates tokens from multiple resized views, introducing external localization priors and a relatively dense token sequence. MSViT~\citep{havtorn2023msvit} learns to choose between coarse and fine tokenization scales across image regions. These approaches demonstrate the value of non-uniform visual processing, but leave an important trade-off unresolved: how can a pretrained ViT preserve its full-image representation while concentrating additional token capacity on task-relevant regions, without explicit part annotations or an expensive attention-based selection pass at inference? This problem is particularly important in Generalized Category Discovery (GCD)~\citep{vaze2022generalized,cao2021open, han2021autonovel, zhao2021novel}, where supervision covers only known categories and the learned representation must also separate unlabeled novel categories. A useful localization strategy must therefore discover transferable fine-grained cues rather than overfit to parts that distinguish only the labeled classes.

To address this trade-off, we propose \textbf{Subtoken Vision Transformer (\ours)} and its \textbf{Attention-based Token Subdivision (ATS)} mechanism. ATS selects the top-$K$ discriminative patches and represents each selected patch with $f\!\times\!f$ subtokens, which are concatenated with all original tokens. Representing an important patch with multiple embeddings exposes its intra-patch spatial structure to the Transformer instead of compressing it into a single vector; retaining the base tokens simultaneously preserves the global image layout. Thus, \ours does not densely increase image resolution or discard background context. It selectively increases representational granularity in informative regions and confines the additional computation to a small subset of patches. Fig.~\ref{fig:intro} contrasts this selective refinement with uniform tokenization and dense multi-scale inputs.

Learning where to subdivide remains difficult because different attention heads capture complementary semantics, while averaging them can dilute localized evidence. We address this difficulty with two training stages that separate exploration from efficient selection. In Stage 1, we randomly sample attention heads to generate diverse subdivision patterns while fine-tuning the ViT, encouraging the representation to remain effective across different potentially discriminative regions. In Stage 2, the Stage-1 backbone is frozen and used as a teacher. We remove the regions selected by each head and measure the resulting feature degradation; the attention map producing the largest degradation provides token-level supervision for a lightweight router. At inference, the teacher branch is removed, and the router directly predicts one deterministic importance map before a single Transformer pass. This converts diverse attention-based exploration into efficient top-$K$ selection without a separate attention forward.

Experimentally, \ours raises average novel-category accuracy from $81.3\%$ to $84.7\%$ over DINOv2 on three fine-grained GCD benchmarks. The efficiency gain is also practical: at $K=1\%$, \ours is only $0.50$ ms above DINOv2, while reducing latency by $73.8\%$ relative to Retina Patch. In summary, our contributions are threefold:

\begin{itemize}[itemsep=1pt, topsep=1pt, leftmargin=*]
    \item We introduce \textbf{\ours} and \textbf{ATS}, a selective tokenization strategy that represents discriminative patches with multiple subtokens while retaining the original token sequence for global context.
    \item We develop a \textbf{two-stage learning strategy} that combines random attention-head exploration with feature-degradation-guided distillation, producing a deterministic single-map router that eliminates the separate attention-based selection pass at inference.
    \item Experiments on three fine-grained GCD benchmarks demonstrate improved recognition of novel categories with controlled computation, together with interpretable, domain-specific subdivision patterns.
\end{itemize}

\section{Related Works}
\label{sec:related_works}


\paragraph{\textbf{Dynamic Token Scaling.}}
Vision Transformers (ViTs), such as CLIP~\citep{radford2021learning}, SigLIP~\citep{zhai2023sigmoid}, and the DINO family~\citep{caron2021emerging, oquab2024dinov}, partition images into non-overlapping tokens and leverage self-attention for feature extraction, contrasting with CNN-based convolutional feature hierarchies. Recent approaches address visual feature extraction through distinct architectural strategies. Swin Transformer~\citep{liu2021swin} and Pyramid vision transformer~\citep{wang2021pyramid} employ multi-scale feature aggregation through shifted windowing, though they introduce computational redundancy while overlooking intra-image heterogeneity. \citet{kim2025sapiensid} introduces Retina Patch with multi-scale tokens concatenation for ViT. These methods demonstrate the importance of spatial adaptation but lack dynamic resource allocation mechanisms for region-specific processing demands. MSViT~\citep{havtorn2023msvit} proposes a mix-scale token using a learnable gating to choose between coarse or fine tokens in every region, but fails to provide flexible choices in adaptive selection~\cite{rao2021dynamicvit, ryoo2021tokenlearner, zhu2026depthagent, yin2022avit, liang2022evit, ma2023dit}. In this work, we propose an Attention-based Token Subdivision (ATS), dynamically subdividing tokens to amplify discriminative features while maintaining pretrained ViT efficiency.

\paragraph{\textbf{Fine-grained Localization.}}
Fine-grained localization pinpoints discriminative features critical for inter-class differentiation, primarily via keypoint detection~\citep{cao2017realtime,fang2022alphapose, yang2021transpose, kim2024keypoint, teepe2022towards} and class activation maps (CAMs). Beyond classification, such localized cues also underpin human recognition systems that fuse face, body, and gait evidence~\citep{liu2025person, zhu2025quality, zhu2026fusionagent, su2026localscore, su2026sapiensid2, su2024open, liu2024farsight, su2025hamobe} and related human-centric analysis and synthesis~\citep{guo2026holistic, xu2026emotag}. However, defining consistent keypoints for generic objects (\eg, industrial products)~\citep{yang2022apt, yu2021ap} remains challenging due to the lack of anatomical priors, risking overemphasis on unimportant features. CAM-based methods~\citep{chowdhury2025prompt, jiang2021layercam, zhou2016learning} generate class-specific saliency maps to identify crucial regions, but are limited in propagating them into feature refinement. To bridge this gap, we propose an attention-driven token selection method that identifies and amplifies discriminative tokens, optimizing fine-grained classification without relying on predefined keypoints.

\paragraph{\textbf{Fine-grained recognition.}}
Fine-grained recognition aims to distinguish subordinate categories that share similar global appearances but differ in localized attributes, such as birds, aircraft, and vehicles~\cite{wah_branson_welinder_perona_belongie_2011, maji2013fine, krause20133d, nilsback2008automated, parkhi2012cats,ren2023chatgpt}. Success therefore depends on preserving subtle, spatially localized evidence rather than relying only on holistic object representations. Part-based models~\cite{zhang2014part}, \eg, constellations~\cite{weber2000unsupervised, fergus2003object}, DPMs~\cite{felzenszwalb2010object, zhang2013deformable}, poselets~\cite{bourdev2009poselets}, and S3N~\cite{ding2019selective}, localize such evidence for fixed classifiers, whereas \ours subdivides selected patches. Recent work leverages multimodal large language models for fine-grained recognition~\cite{chen2023atm, jiang2024mpfgvc, sun2023fgvpl, wei2024cascadevlm, he2025finedefics, xie2025fgclip, xiao2025flair, Zhu_2026_CVPR}. Generalized Category Discovery (GCD) is a particularly challenging fine-grained recognition setting because it tackles the task of jointly recognizing known classes and discovering novel categories in unlabeled data, as formalized by \citet{vaze2022generalized} and \citet{cao2021open}. GCD extends novel category discovery (NCD)~\citep{han2021autonovel, zhao2021novel, zhong2021openmix, fini2021unified} by requiring models to simultaneously leverage labeled data and partition unlabeled novel subcategories. We adapt the GCD task to evaluate the fine-grained capturing capability of \ours and investigate two aspects: (1) Which object tokens are pivotal for fine-grained differentiation? and (2) Can targeted token subdivision enhance feature discernibility?

\section{Methods} \label{sec:methods}

\subsection{Preliminary}

\paragraph{Image Tokenization.}
Given an image tensor $ x \in \mathbb{R}^{C \times H \times W} $, ViTs partition $ x $ into $ N = \frac{H}{P} \times \frac{W}{P} $ non-overlapping patches, where $P$ is the patch size. All patches are stacked into an image patch sequence $ I \in \mathbb{R}^{N \times C \times P \times P} $. $I$ is flattened in its last three dimensions, and a learnable projection matrix $ \mathbf{W_p} \in \mathbb{R}^{C P^2 \times D} $ (\ie, patch embedding layer) transforms $ I $ into patch embeddings, combined with positional embeddings $ E_{\text{pos}} \in \mathbb{R}^{N \times D}$ to form the input sequence $ z_0 \in \mathbb{R}^{N \times D} $.
\paragraph{Attention Map.}
For input tokens $ \mathbf{z}_l \in \mathbb{R}^{N \times D} $ at transformer block $ l $, the self-attention mechanism computes queries $ \mathbf{Q} = \mathbf{z}_l \mathbf{W}_Q $, keys $ \mathbf{K} = \mathbf{z}_l \mathbf{W}_K $, and values $ \mathbf{V} = \mathbf{z}_l \mathbf{W}_V $, where $ \mathbf{W}_Q, \mathbf{W}_K, \mathbf{W}_V \in \mathbb{R}^{D \times d_k} $ are learnable weights to generate attention map $ \mathbf{A}_l \in \mathbb{R}^{N \times N} $. 
Conventionally, the attention map from the CLS token $ \mathbf{A}_{l, \text{CLS}} \in \mathbb{R}^{N} $  represents the global attention for the model. Each element $ \mathbf{A}_{l, \text{CLS}}{(j)} $ quantifies how much the CLS token attends to token $ j $, reflecting the model’s focus on discriminative regions (\eg, object parts). Higher values indicate stronger semantic relevance between tokens. ViTs exhibit hierarchical attention patterns across layers, with deeper blocks increasingly focusing on semantically fine-grained details, while shallow layers prioritize low-level textures or global context~\citep{dosovitskiy2020image, chefer2021transformer}. With multi-head self-attention, the CLS attention maps of all $N_{\text{head}}$ heads are stacked as $\hat{\mathbf{A}}_{l, \text{CLS}} \in \mathbb{R}^{N_{\text{head}} \times N}$.

\begin{figure}[t!]
    \centering
    \includegraphics[width=0.9\linewidth]{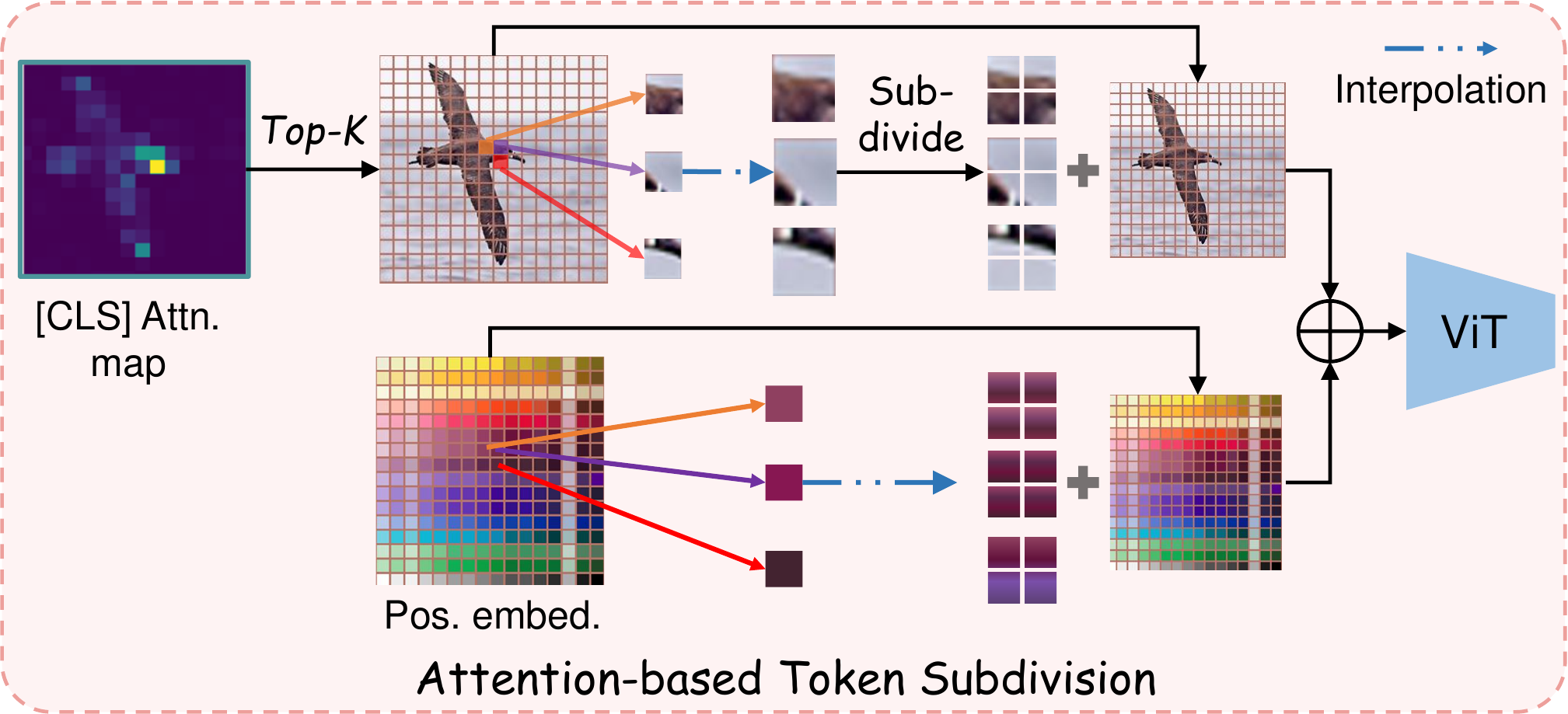}
    \caption{Overview of ATS. We select the top-$K$ patches based on the attention map for subdivision. We select the neighboring positional embedding for interpolation purposes.}
    \label{fig:ats}
\end{figure}

\subsection{Attention-based Token Subdivision (ATS)}

Defining consistent keypoints for generic objects remains challenging due to the lack of anatomical priors, risking overemphasis on non-discriminative features. We introduce ATS to leverage knowledge from attention maps. 

\paragraph{Image Subdivision.} As shown in Fig.~\ref{fig:ats}, given an image patch sequence $I$ and its attention map $\mathbf{A}^h_{l, \text{CLS}}$ from head $h$, we locate the top $K$ patch set $T \in \mathbb{R}^{K \times (C \times P \times P)}$ sorted by the attention scores. For each patch, we generate $f\!\times\!f$ subtokens. We first bilinearly upscale each patch by
\begin{align}
    \hat{T}_i\!=\!\operatorname{Upscale}(T_i, \text{factor}\!=\!f), \quad \hat{T}_i \in \mathbb{R}^{C \times (fP) \times (fP)}.
\end{align}
Then we divide the interpolated patch into an $f\times f$ grid of smaller patches indexed by $m,n \in \{0, \dots, f-1\}$. Specifically, one subtoken is defined as:
\begin{align}
    \begin{aligned}
        \hat{T}_i^{(m,n)} &= \hat{T}_i[:, mP\!:\!(m\!+\!1)P, nP\!:\!(n\!+\!1)P],
    \end{aligned}
\end{align}
where all subtokens are denoted as:
\begin{align}
    T' = \bigcup_{i=1}^{K} \bigcup_{m=0}^{f-1} \bigcup_{n=0}^{f-1} \hat{T}_i^{(m,n)}  \in \mathbb{R}^{Kf^2 \times (C \times P \times P)}. 
\end{align}
Simply put, subtokens are created by upscaling and dividing each patch.
$T'$ is concatenated with $I$ into the input sequence $I^{\text{in}}$ of $N+Kf^2$ patches, embedded by the shared projection $\mathbf{W_p}$. Retaining the parent patches $T$ lets $T$ and $T'$ provide coarse and fine views of the same region.

\begin{figure*}[t!]
    \centering
    \includegraphics[width=0.9\linewidth]{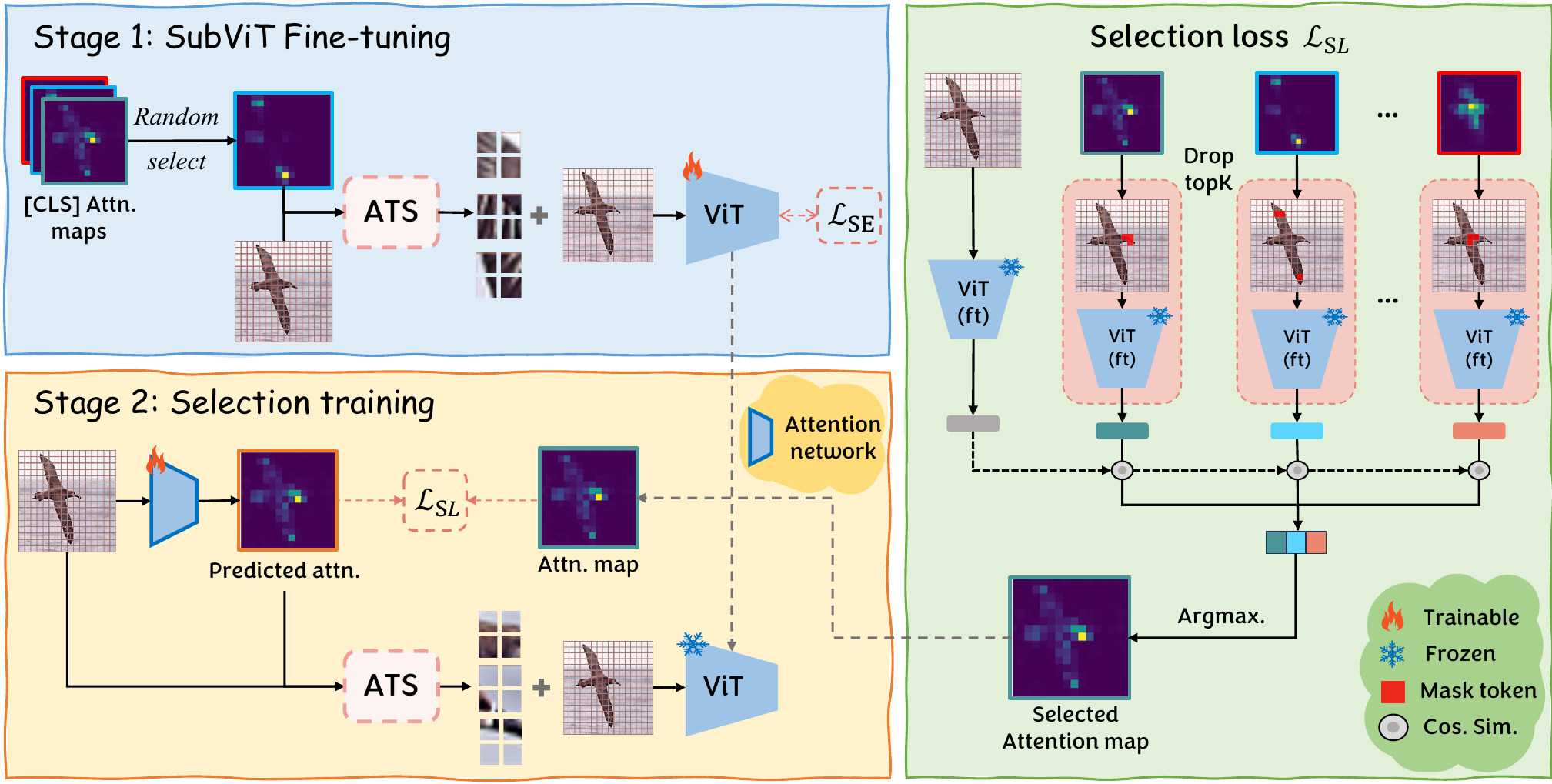}
    \caption{Two-Stage Fine-tuning Framework. Stage 1 performs attention shift adaptation through the randomized selection of attention maps for \ours fine-tuning. Stage 2 introduces a lightweight attention network that predicts the attention scores for each image token using selection loss (right), computed as feature degradation distance with maximum-distance head indices as pseudo-labels for selection prediction.}
    \label{fig:training}
\end{figure*}

\paragraph{Positional Embedding Interpolation.} 
Positional embeddings (PE) are also subdivided in a similar fashion to image patches. However, while image patches can be upscaled due to their spatial resolution, a positional embedding at a particular location is a vector and cannot be directly upscaled without remaining unchanged across the newly introduced positions. 

To address this, we expand each positional embedding into a spatial structure by using its neighboring positional embeddings. Given an original positional embedding $E_{\text{pos}, i}$, we generate an upscaled embedding patch by interpolation:
\begin{align}
    \hat{E}_{\text{pos}, i} = \operatorname{Upscale}(\{E_{\text{pos}, j} \mid j \in \mathcal{N}(i) \}, f),
\end{align}
where $\mathcal{N}(i)$ denotes $E_{\text{pos}, i}$ and its eight adjacent positional embeddings, and $\operatorname{Upscale}$ is bicubic interpolation.
After interpolation, we divide $\hat{E}_{\text{pos}, i}$ into smaller positional embeddings in an $f \times f$ grid, similar to image patch subdivision:
\begin{align}
    \hat{E}_{\text{pos}, i}^{(m,n)} = \hat{E}_{\text{pos}, i}[:, m, n], 
\end{align}
for all $m,n \in \{0, \dots, f-1\}$.
We then gather all subdivided positional embeddings into the final expanded set:
\begin{align}
    E_{\text{pos}}^{T'} = \bigcup_{i=1}^{K} \bigcup_{m=0}^{f-1} \bigcup_{n=0}^{f-1} \hat{E}_{\text{pos}, i}^{(m,n)}  \in \mathbb{R}^{K f^2 \times D}.
\end{align}
Finally, the expanded positional embeddings are concatenated with the original positional embeddings to maintain spatial consistency in the input sequence:
\begin{align}
    z_0^{\text{new}} = \mathbf{W_p} I^{\text{in}} + \operatorname{Concat}(E_{\text{pos}}, E_{\text{pos}}^{T'}).
\end{align}
This ensures that the new PEs retain meaningful spatial structure.

\subsection{Two-stage Fine-Tuning} \label{subsec:2stage_finetuning}

A key challenge in subtoken division is to obtain informative and diverse subdivision regions without requiring an additional Transformer forward pass at inference. Averaging all attention heads repeatedly emphasizes similar regions, whereas using a fixed head ignores the complementary semantics learned by multi-head attention. We therefore adopt two stages: randomized attention-guided fine-tuning, then distance-selected attention-map distillation.

\paragraph{\ours Fine-Tuning (Stage 1).}  
Stage 1 adapts the pretrained ViT to subdivided token sequences whose selected regions vary during training. In each forward pass, we uniformly sample one head $h$ and select the top-$K$ patches from its CLS-to-patch attention map $\mathbf{A}^h_{L,\text{CLS}}$ of the last block $L$, computed on the base tokens. The model is then fine-tuned on the resulting $N+Kf^2$ tokens with the SelEx objective. Sampling heads exposes the model to complementary subdivision patterns instead of a single fixed region, while retaining $I$ preserves the global layout learned during pretraining. The adapted backbone serves as the teacher in Stage 2.

\paragraph{Distance-guided Token Selection Training (Stage 2).}
We freeze the Stage-1 ViT and use it as a teacher during Stage 2. For each image, the teacher first produces the original feature $x_{\mathrm{ori}}$ and the CLS-to-patch attention map $A^h\in\mathbb{R}^{N}$ for every head $h$. We then remove the top-$K$ patches indicated by each head and forward the remaining tokens through the frozen teacher to obtain the degraded feature $x_{\mathrm{drop}}^h$. Following the selection-loss motivation, the importance label of head $h$ is its feature-degradation distance
\begin{align}
    d^h = \left\lVert x_{\mathrm{ori}}-x_{\mathrm{drop}}^h\right\rVert_2,
    \qquad h^*=\operatorname*{arg\,max}_{h} d^h.
\end{align}
The largest distance indicates that removing the corresponding regions causes the greatest representation degradation. We therefore use the selected map $A^{h^*}$ as token-level supervision for a lightweight attention network $R$. Given the base patch embeddings $z\in\mathbb{R}^{N\times D}$, $R$ predicts a single score map $\hat{A}=R(z)\in\mathbb{R}^{N}$ rather than reproducing all attention heads. Specifically, $R$ normalizes $z$, projects it from $768$ to $192$ channels, and reshapes it into the $16\times16$ patch grid; three depthwise residual blocks ($3\times3$ depthwise and $1\times1$ pointwise convolutions with BatchNorm and GELU) aggregate local context, a global channel gate computed from mean-pooled tokens injects image-level context, and a $1\times1$ convolution outputs one score per patch. $R$ has only ${\sim}0.34$M parameters (${\sim}0.4\%$ of the ViT).

We retain the map, ranking, and top-$K$ objectives for router distillation. With temperature $\tau$, the map loss matches the score distributions,
\begin{align}
    \mathcal{L}_{\mathrm{map}} = \tau^2\operatorname{KL}\!\left(
    \operatorname{Softmax}(\log A^{h^*}/\tau)\,\|\,
    \operatorname{Softmax}(\hat{A}/\tau)\right).
\end{align}
The ranking loss maximizes the Pearson correlation of router scores with teacher ranks, and the top-$K$ loss is a cross-entropy on $\mathcal{T}{=}\operatorname{TopK}(A^{h^*})$:
\begin{align}
    \mathcal{L}_{\mathrm{rank}} = 1-\operatorname{corr}\big(\hat{A}, \operatorname{rank}(A^{h^*})\big), \quad
    \mathcal{L}_{\mathrm{topK}} = -\frac{1}{K}\sum\nolimits_{i\in\mathcal{T}}\log\operatorname{Softmax}(\hat{A})_i.
\end{align}
The Stage-2 objective is $\mathcal{L}_{\mathrm{router}}=\lambda_{\mathrm{map}}\mathcal{L}_{\mathrm{map}}+\lambda_{\mathrm{rank}}\mathcal{L}_{\mathrm{rank}}+\lambda_{\mathrm{topK}}\mathcal{L}_{\mathrm{topK}}$.
The teacher's attention and degraded-feature forwards are required only during Stage-2 training. At inference, the frozen teacher branch is removed: the router directly predicts one deterministic score map, whose top-$K$ patches are subdivided before a single ViT forward.

\section{Experiments} \label{sec:experiments}
\subsection{Experimental Setup}

\paragraph{Datasets and GCD Protocol.} We evaluate \ours on three fine-grained datasets: \mbox{CUB-200}~\citep{wah_branson_welinder_perona_belongie_2011}, FGVC-Aircraft (Aircraft)~\citep{maji2013fine}, and Stanford-Cars (SCars)~\citep{krause20133d}. Following the standard GCD protocol~\citep{vaze2022generalized} adopted by SelEx, half of the categories are known; labeled data are sampled from known-category training images, while the remaining known-category images and all novel-category images are unlabeled and must be clustered without novel labels (labeled/unlabeled images: 1.5K/4.5K for CUB, 3.4K/6.6K for Aircraft, and 2.0K/6.1K for SCars). We cluster all features with balanced semi-supervised $k$-means (BSSK)~\citep{rastegar2024selex} and report \textit{All}/\textit{Known}/\textit{Novel} accuracy on the unlabeled set; Hungarian matching~\citep{wright1990speeding} is used only for scoring, without any out-of-distribution heuristic.

\paragraph{Baseline Setup.} We compare \ours with classic ViT~\citep{dosovitskiy2020image, oquab2024dinov}, Retina Patch~\citep{kim2025sapiensid} and MSViT~\citep{havtorn2023msvit}, which apply different strategies towards image tokenization. Note that the ViT baseline fine-tuned with the same loss corresponds to the full SelEx~\citep{rastegar2024selex} method, a state-of-the-art GCD approach. We use CapeX~\citep{rusanovsky2024capex} as the keypoint predictor for Retina Patch. For a fair comparison, all methods share the GCD split, DINO backbone, and SelEx loss and hyperparameters, so that accuracy differences stem from tokenization; fixed-classifier FGVC methods are thus not compared quantitatively.

\paragraph{Implementation Details.} We use DINOv2~\citep{oquab2024dinov} pretrained on LVD-142M and DINOv1~\citep{caron2021emerging} pretrained on ImageNet-1K~\citep{deng2009imagenet}. We use the batch size of $128$ for training and set the same loss hyperparameters as SelEx. We fine-tune the last two blocks of the pretrained ViTs. We select $K$ per dataset, setting $K=10\%$ (\ie, top 10\% of image tokens) for CUB, $K=2\%$ for Aircraft, and $K=1\%$ for SCars (Sec.~\ref{subsec:ablation}). We use $f=2$ and $lr=0.1$ for Stage-1 ViT fine-tuning. In Stage 2, the ViT is frozen, and the single-map router is trained for 30 epochs using AdamW with $lr=10^{-4}$. We use Euclidean feature-degradation distance, distillation temperature $\tau=0.5$, and set $\lambda_{\mathrm{map}}=\lambda_{\mathrm{rank}}=\lambda_{\mathrm{topK}}=1$.

\subsection{Comparison with Tokenization Baselines}

\begin{table*}[t!]
\tabcolsep=0.1cm
\caption{\textbf{Comparison with tokenization baselines on fine-grained GCD.} \ours achieves the best \textit{All} and \textit{Novel} accuracy on all datasets, with the largest gains on \textit{Novel}. Bold and underlined numbers indicate the best and second-best accuracies, respectively. $^*$: reported from~\cite{rastegar2024selex}.}
\label{tab:fine-grained_performance}
\centering
\resizebox{\linewidth}{!}{
\begin{tabular}{c|l|ccc|ccc|ccc|ccc}
\toprule
& \multirow{3}{*}{\textbf{Method}} &\multicolumn{3}{c}{\textbf{CUB-200}}& \multicolumn{3}{c}{\textbf{FGVC-Aircraft}}&\multicolumn{3}{c}{\textbf{Stanford-Cars}}&\multicolumn{3}{c}{\textbf{Average}}  \\ 
\cmidrule(lr){3-5} \cmidrule(lr){6-8} \cmidrule(lr){9-11} \cmidrule(lr){12-14}
& &All&Known&Novel&All&Known&Novel&All&Known&Novel&All& Known&Novel\\
\midrule
\multirow{4}{*}{\rotatebox{90}{DINOv1}}
& ViT*~\cite{dosovitskiy2020image} & \underline{73.6} & \underline{75.3} & \underline{72.8} & \underline{57.1} & \bf 64.7 & \underline{53.3} & \underline{58.5}& \underline{75.6} & \underline{50.3} &\underline{63.0}&\underline{71.9}&\underline{58.8}\\
&MSViT~\cite{havtorn2023msvit} & 73.5 & 74.9 & \underline{72.8} & 55.6 & \underline{64.5} & 51.2 & 52.6 & 74.0 & 42.3 & 60.6 & 71.1 & 55.4 \\
&Retina Patch~\cite{kim2025sapiensid} & 71.6 & 73.5 & 70.7 & 52.9 & 57.7 & 50.5 & 52.0 & 72.9 & 41.9 & 58.8 & 68.0 & 54.4 \\  
& \cellcolor{green!20}\textbf{\ours (Ours)} 
   & \cellcolor{green!20}\bf 75.7 & \cellcolor{green!20}\bf 76.4 & \cellcolor{green!20}\bf 75.4 
   & \cellcolor{green!20}\bf 57.5 & \cellcolor{green!20}64.1 & \cellcolor{green!20}\bf 54.2 
   & \cellcolor{green!20}\bf 59.3 & \cellcolor{green!20}\bf 76.0 & \cellcolor{green!20}\bf 52.1 
   & \cellcolor{green!20}\bf 64.2 & \cellcolor{green!20}\bf 72.2 & \cellcolor{green!20}\bf 60.6 \\

\midrule
\multirow{4}{*}{\rotatebox{90}{DINOv2}}
&ViT*~\cite{dosovitskiy2020image} & 87.4 &85.1&88.5&79.8& \bf{82.3}&78.6& \underline{82.2} &93.7& \underline{76.7} &83.1& \underline{87.0} &81.3 \\
&MSViT~\cite{havtorn2023msvit} & \underline{88.3} & 85.7 & \underline{90.0} & \underline{79.9} & 79.6 & \underline{80.0} & 81.9 & 93.1& 76.5 & \underline{83.4} & 86.1 & \underline{82.2} \\
&Retina Patch~\citep{kim2025sapiensid}& 87.8 & \underline{86.1} & 88.8 & 73.2 & 74.0 & 72.8 & 80.4 & \underline{93.8} & 73.9 & 80.5 & 84.6 & 78.5 \\
& \cellcolor{green!20}\textbf{\ours (Ours)} 
    & \cellcolor{green!20}\bf{91.8} & \cellcolor{green!20}\bf{86.3} & \cellcolor{green!20}\bf{94.6} 
    & \cellcolor{green!20}\bf{80.8} & \cellcolor{green!20}\underline{81.3} & \cellcolor{green!20}\bf{80.5} 
    & \cellcolor{green!20}\bf{83.8} & \cellcolor{green!20}\bf{94.9} & \cellcolor{green!20}\bf{78.5} 
    & \cellcolor{green!20}\bf{85.5} & \cellcolor{green!20}\bf{87.5} & \cellcolor{green!20}\bf{84.7} \\
\bottomrule
\end{tabular}
}
\end{table*}

\paragraph{Fine-grained Image Classification.} Tab.~\ref{tab:fine-grained_performance} compares \ours with tokenization baselines on three fine-grained datasets. \ours achieves the best \textit{All} and \textit{Novel} accuracy on every dataset with both backbones. Compared with Retina Patch, it attains higher accuracy with far fewer tokens, suggesting that dense multi-view tokens add redundant evidence that fits pretrained ViTs poorly. MSViT gains little with DINOv2 and degrades with DINOv1, suggesting that choosing a finer scale per region adds limited fine-grained information. In contrast, \ours adds subtokens only to selected patches while keeping the loss function and backbone architecture unchanged. With DINOv2, it improves average \textit{Novel} accuracy by $3.4\%$ while keeping \textit{Known} accuracy comparable ($+0.5\%$). The gains are largest on CUB ($+4.4$ \textit{All}, $+6.1$ \textit{Novel}) and smaller on Aircraft ($+1.0$/$+1.9$) and SCars ($+1.6$/$+1.8$), consistent with the object-scale trend in Sec.~\ref{subsec:ablation}: birds occupy a small image region and benefit most from local refinement.

\begin{table}[t!]
\centering
\tabcolsep=0.1cm
\caption{\textbf{Comparison of inference efficiency and resource usage.} All measurements use batch size 128 on an NVIDIA A6000 GPU and are amortized per image. ``Two-pass'' uses an initial attention forward for region selection followed by a refinement forward, whereas our router directly predicts a single importance map.}
\vspace{0.3em}
\label{tab:efficiency}
\resizebox{0.9\linewidth}{!}{
    \begin{tabular}{l|c|ccc}
    \toprule
    \textbf{Model} & \textbf{\# Tokens} & \textbf{Latency (ms)} & \textbf{Memory (MB)} & \textbf{FLOPs (G)} \\
    \midrule
    DINOv2~\cite{oquab2024dinov}                                  & 256 & 3.07 & 8.42 & 22.30 \\
    MSViT~\cite{havtorn2023msvit}                                   & 256 & 3.30 & 9.18 & 22.34 \\
    Retina Patch~\cite{kim2025sapiensid}                            & 711 & 13.63 & 35.35 & 82.68 \\
    \midrule
    \ours Two-pass ($K=1\%$)                & 264 & 6.68 & 16.76 & 45.93 \\
    \textbf{\ours Router ($K=1\%$)}         & 264 & 3.57 & 10.24 & 23.06 \\
    \midrule
    \ours Two-pass ($K=2\%$)                & 276 & 6.85 & 16.75 & 46.96 \\
    \textbf{\ours Router ($K=2\%$)}         & 276 & 3.74 & 10.71 & 24.08 \\
    \midrule
    \ours Two-pass ($K=10\%$)               & 356 & 8.31 & 16.75 & 53.80 \\
    \textbf{\ours Router ($K=10\%$)}        & 356 & 5.21 & 13.76 & 30.92 \\
    \bottomrule
    \end{tabular}
}
\vspace{-1em}
\end{table}

\begin{table}[t!]
\caption{\textbf{Inference-time pruning of low-scored base tokens on CUB.} All subtokens are kept without retraining; $\Delta$ is the accuracy change relative to no pruning.}
\label{tab:bottom_pruning}
\centering
\tabcolsep=0.12cm
\resizebox{0.8\linewidth}{!}{
\begin{tabular}{c|c|cc|ccc}
\toprule
\textbf{Pruned} & \textbf{\# Tokens} & \textbf{Tokens} $\downarrow$ & \textbf{FLOPs} $\downarrow$ & $\Delta$\textit{All} & $\Delta$\textit{Known} & $\Delta$\textit{Novel} \\
\midrule
$0\%$  & 356 & $0.0\%$  & $0.0\%$  & $0.00$  & $0.00$  & $0.00$  \\
$5\%$  & 344 & $3.4\%$  & $3.3\%$  & $+0.02$ & $+0.07$ & $0.00$  \\
$10\%$ & 331 & $7.0\%$  & $6.9\%$  & $-0.63$ & $0.00$  & $-0.93$ \\
$20\%$ & 305 & $14.3\%$ & $14.0\%$ & $-0.67$ & $0.00$  & $-1.00$ \\
$30\%$ & 280 & $21.3\%$ & $20.9\%$ & $-0.92$ & $-0.27$ & $-1.23$ \\
\bottomrule
\end{tabular}}
\vspace{-1em}
\end{table}

\paragraph{Inference Efficiency.} Tab.~\ref{tab:efficiency} separates the cost of fine-grained token processing from that of selecting subdivision regions. The earlier two-pass variant first extracts backbone attention to determine important regions and then processes the resulting token sequence, repeating a substantial portion of the backbone computation. In contrast, the single-map router predicts token scores directly from base patch embeddings and requires only one backbone pass. Under identical token budgets, the router reduces latency by $37.3$--$46.6\%$, peak memory by $17.9$--$38.9\%$, and FLOPs by $42.5$--$49.8\%$ across the three $K$ settings. At $K=1\%$, \ours takes $3.57$ ms per image, only $0.50$ ms ($16.3\%$) slower than the single-stream DINOv2 baseline, with just $3.4\%$ additional FLOPs. In comparison, it reduces latency by $73.8\%$ relative to Retina Patch while retaining targeted subtoken refinement. Increasing $K$ provides a controllable trade-off between local detail and computation. Even at $K=10\%$, the router reduces latency, memory, and FLOPs relative to Retina Patch by $61.8\%$, $61.1\%$, and $62.6\%$, respectively, because it refines selected regions instead of concatenating several dense multi-scale views. Router scores also permit inference-time pruning of low-scored base tokens without retraining (Tab.~\ref{tab:bottom_pruning}): pruning $5\%$ cuts FLOPs by $3.3\%$ with unchanged accuracy, and larger ratios trade accuracy for FLOPs gradually. Notably, \textit{Known} accuracy is unaffected up to $20\%$ pruning, whereas \textit{Novel} accuracy drops by about one point, suggesting that low-scored tokens still carry context useful for discovering novel categories.

\subsection{Ablation Studies} \label{subsec:ablation}

\begin{figure}[t!]
\centering
\includegraphics[width=1\linewidth]{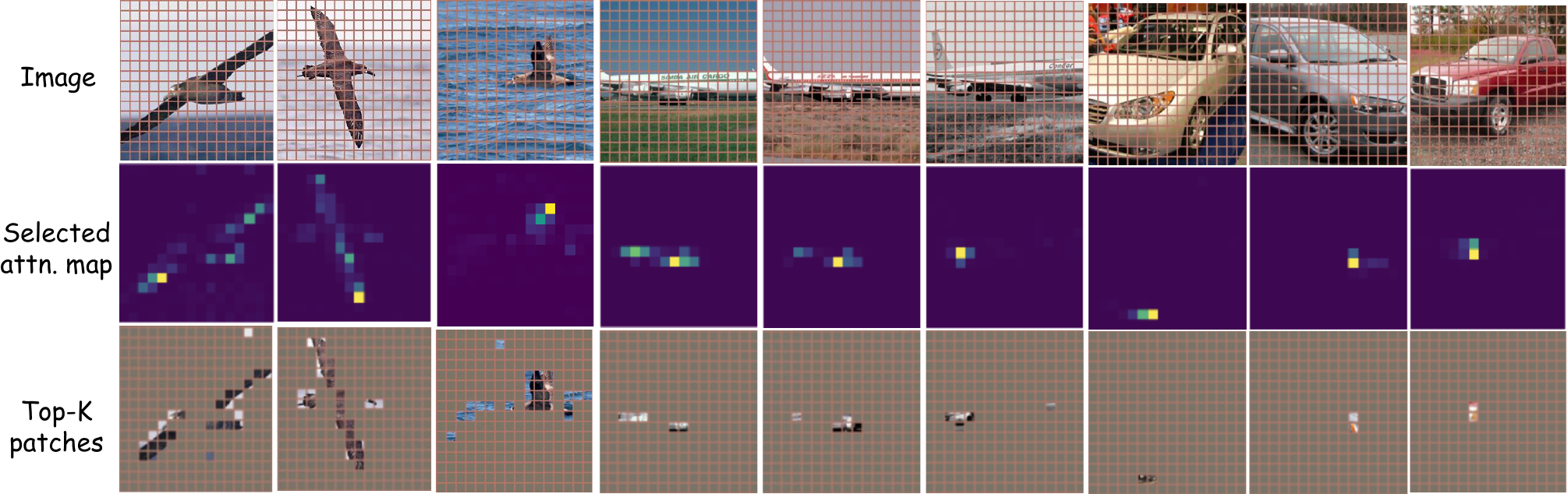}
\caption{\textbf{Visualization of attention map selection.} $K=10\%$ for CUB, $2\%$ for Aircraft, and $1\%$ for SCars. \ours demonstrates consistent semantic region selection across instances through the token subdivision. Zoom in for a better effect.}
\label{fig:selection_visualization}
\end{figure}

\paragraph{Effects of Token Selection Methods.} We compare our token selection strategy against two sampling strategies: (1) random sampling of $K$ tokens and (2) top-$K$ selection from the attention map averaged across heads. As shown in Tab.~\ref{tab:ablation_selection}, random subdivision decreases \textit{All} accuracy from $87.4\%$ to $82.4\%$, indicating that merely increasing the token count does not guarantee better representations. Averaged attention improves the baseline, particularly on novel categories, but still treats complementary head semantics uniformly. Our learned selection further improves \textit{All} and \textit{Novel} accuracy over averaged attention by $2.8\%$ and $3.4\%$, respectively. This gap supports the importance of selecting semantically informative subdivision targets rather than uniformly aggregating or randomly sampling them. Fig.~\ref{fig:selection_visualization} provides corresponding qualitative evidence through the consistent localization of discriminative regions.

\begin{table}[t!]
\caption{\textbf{Ablation studies of \ours on CUB.} (a) Effects of token selection methods; (b) Effects of $f$ and $K$; (c) Effects of Masking tokens.}
\label{tab:combined_ablation_all}
\vspace{-1em}
\tabcolsep=0.1cm
\centering
\begin{subtable}[t]{0.38\textwidth}
    \centering
    \caption{\textbf{Token selection methods.}}
    \resizebox{\linewidth}{!}{
    \begin{tabular}{l|c|ccc}
      \toprule
      \textbf{Method} & \# Tokens & All & Known & Novel\\
      \midrule
      DINOv2 & 256 & 87.4 & 85.1 & 88.5 \\
      Random & 356 & 82.4 & 85.2 & 81.0 \\ 
      Average & 356 & 89.0 & 84.6 & 91.2 \\ 
      \rowcolor{green!20} \textbf{Ours} & 356 & 91.8 & 86.3 & 94.6 \\
      \bottomrule
    \end{tabular}
    }
    \label{tab:ablation_selection}
\end{subtable}
\hfill
\begin{subtable}[t]{0.29\textwidth}
    \centering
    \caption{\textbf{Effects of $f$ and $K$.}}
    \resizebox{\linewidth}{!}{
    \begin{tabular}{cc|ccc}
      \toprule
      $f$ & $K (\%)$  & All & Known & Novel\\
      \midrule
      2 & 1 & 90.3 & 84.0 & 93.5 \\
      2 & 10 & 91.8 & 86.3 & 94.6 \\
      3 & 1 & 90.4 & 84.2 & 93.6 \\
      3 & 10 & 90.7 & 85.9 & 93.3 \\
      \bottomrule
    \end{tabular}
    }
    \label{tab:ablation_params}
\end{subtable}
\hfill
\begin{subtable}[t]{0.29\textwidth}
    \centering
    \caption{\textbf{Masking tokens.}}
    \resizebox{\linewidth}{!}{
    \begin{tabular}{l|ccc}
      \toprule
      \textbf{Method} & All & Known & Novel\\
      \midrule
      \rowcolor{green!20} Ours & 91.8 & 86.3 & 94.6 \\
      \cmidrule{1-4}
      w/o $T$ & 90.9 & 84.5 & 94.1 \\
      w/o $T'$& 89.7 & 84.6 & 92.4 \\
      w/o $T + T'$& 68.6 & 65.1 & 70.3 \\
      \bottomrule
    \end{tabular}
    }
    \label{tab:ablation_drop}
\end{subtable}
\end{table}

\begin{table}[t!]
\caption{\textbf{Two-stage fine-tuning performance comparison on Aircraft and SCars.} The results indicate the effectiveness of the proposed two-stage training.}
\label{tab:ablation_stage_performance}
\vspace{-1em}
\centering
\tabcolsep=0.1cm
\begin{subtable}[t]{0.48\textwidth}
    \centering
    \caption{Stage 1 and 2 Comparison on Aircraft.}
    \label{tab:ablation_stage_aircraft}
    \resizebox{\linewidth}{!}{
      \begin{tabular}{l|ccc}
      \toprule
      \textbf{Method} & All & Known & Novel\\
      \midrule
      DINOv2 & 79.8 & 82.3 & 78.6 \\
      \cmidrule{1-4}
      Stage 1 & 80.0 & 81.0 & 79.3 \\
      Stage 2 & 80.8 (\textcolor{green}{$+0.8\%$}) & 81.3 (\textcolor{green}{$+0.3\%$}) & 80.5 (\textcolor{green}{$+0.8\%$}) \\
      \bottomrule
      \end{tabular}
    }
\end{subtable}
\hfill
\begin{subtable}[t]{0.48\textwidth}
    \centering
    \caption{Stage 1 and 2 Comparison on SCars.}
    \label{tab:ablation_stage_scars}
    \resizebox{\linewidth}{!}{
      \begin{tabular}{l|ccc}
      \toprule
      \textbf{Method} & All & Known & Novel\\
      \midrule
      DINOv2 & 82.2 & 93.7 & 76.7 \\
      \cmidrule{1-4}
      Stage 1 & 83.0 & 93.6 & 78.0 \\
      Stage 2 & 83.8 (\textcolor{green}{$+0.8\%$}) & 94.9 (\textcolor{green}{$+1.1\%$}) & 78.5 (\textcolor{green}{$+0.5\%$}) \\
      \bottomrule
      \end{tabular}
    }
\end{subtable}
\end{table}

\paragraph{Effects of Hyperparameters.} Tab.~\ref{tab:ablation_params} presents the effects of scale factor $f$ and top-$K$ selection of \ours. At $K=1\%$, $f=2$ and $f=3$ perform similarly, suggesting that either subdivision scale is sufficient when only a few highly ranked regions are refined. The difference becomes clearer at $K=10\%$: increasing $f$ from $2$ to $3$ reduces \textit{All} accuracy from $91.8\%$ to $90.7\%$, suggesting that aggressively magnifying many regions can introduce overlapping or redundant local evidence. The best setting, $f=2$ and $K=10\%$, balances local resolution with token diversity on CUB. On Aircraft and SCars, smaller budgets are preferable: with $f=2$, increasing $K$ from $1\%$ to $5\%$ and $10\%$ lowers \textit{All} by $0.5$ and $2.1$ points on Aircraft and by $1.9$ and $2.1$ points on SCars. We attribute this trend to object scale: birds occupy a smaller portion of CUB images and benefit from refining more regions, whereas larger aircraft and cars mainly gain redundant detail from additional regions.

\begin{figure*}[t!]
    \centering
    \includegraphics[width=\linewidth]{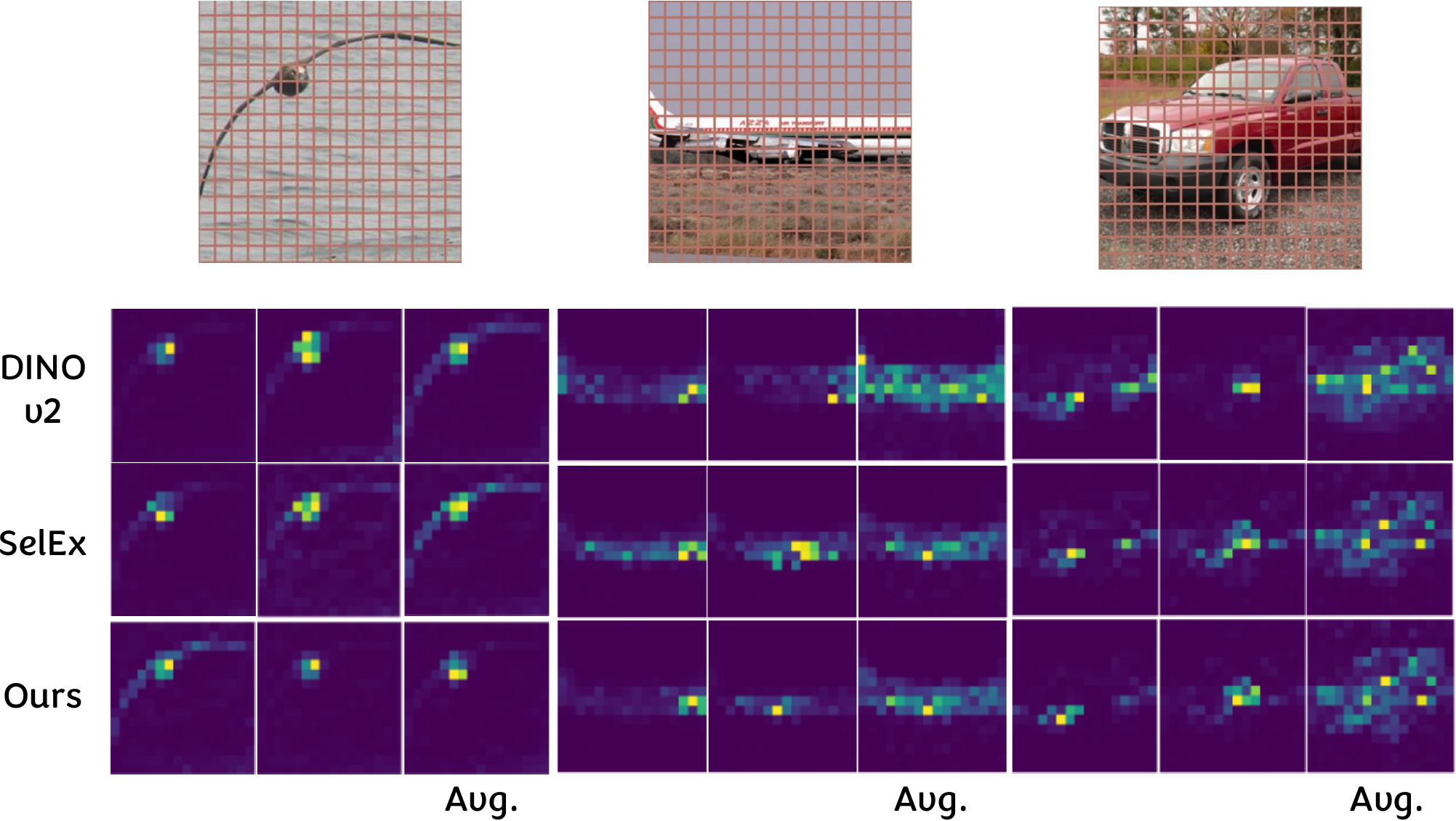}
    \caption{\textbf{Attention map comparison of pretrained DINOv2, SelEx, and \ours on CUB, Aircraft, and SCars.} For each example, we show the CLS attention of representative heads in the last layer and the head-averaged map (Avg.). \ours concentrates attention on fine-grained object regions with more compact and peaked responses, whereas the baselines spread attention over background context.}
    \label{fig:attnmap_comparison}
\end{figure*}

\paragraph{Effects of Token Masking.} Tab.~\ref{tab:ablation_drop} masks the selected parent tokens $T$, the subtokens $T'$, or both at inference. Removing all subtokens $T'$ lowers \textit{All}\slash\textit{Known}\slash\textit{Novel} from $91.8/\allowbreak 86.3/\allowbreak 94.6$ to $89.7/\allowbreak 84.6/\allowbreak 92.4$. This variant still exceeds DINOv2 ($87.4/\allowbreak 85.1/\allowbreak 88.5$) by $2.3/3.9$ points on \textit{All}/\textit{Novel} with comparable \textit{Known}, indicating that part of the gain comes from Stage-1 adaptation, whereas retaining $T'$ adds a further $2.1/2.2$ points. Masking $T$ instead lowers \textit{All} by only $0.9$ points, while masking both $T$ and $T'$ reduces it to $68.6$ because the selected regions are then absent from the input. Thus, $T$ preserves the coarse appearance of selected regions and $T'$ supplies their fine detail.

\paragraph{Performance Comparison of Stage 1 and 2.} Tab.~\ref{tab:ablation_stage_performance} isolates the contribution of the two training stages. Stage 1 already improves novel-category accuracy over DINOv2 by $0.7\%$ on Aircraft and $1.3\%$ on SCars, showing that exposure to diverse subdivisions strengthens the underlying representation. Stage 2 then raises \textit{All} accuracy by a further $0.8\%$ on both datasets, with gains in both known and novel subsets. Thus, the router does more than replace the costly two-pass selection procedure: its distilled supervision preserves the useful attention preferences learned in Stage 1 while converting them into a deterministic and efficient selection rule.

\begin{figure*}[t]
    \centering
    \includegraphics[width=\linewidth]{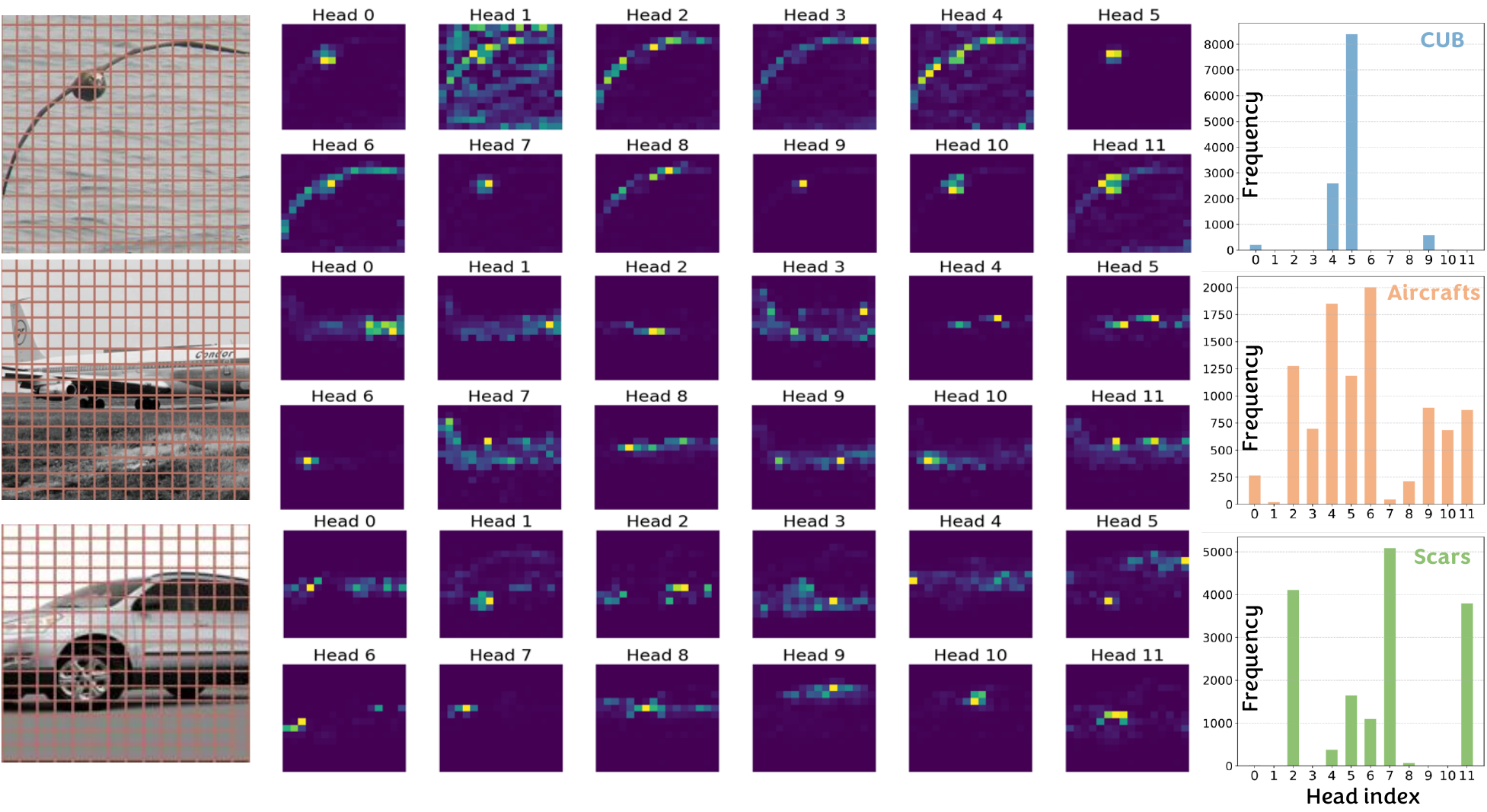}
    \caption{\textbf{Distributions of head selection frequency.} The head selection distribution pattern implies the varying importance of different heads for the target datasets. We select one example from each dataset to visualize the attention pattern of each head in the last layer of DINOv2.}
    \label{fig:head_distribution_visualization}
\vspace{-.5em}
\end{figure*}

\begin{wraptable}{r}{0.40\linewidth}
\vspace{-1.15em}
\centering
\caption{\textbf{Subtoken PE (CUB, Stage 1).}}
\label{tab:ablation_pe}
\tabcolsep=0.1cm
\resizebox{\linewidth}{!}{
\begin{tabular}{l|ccc}
\toprule
\textbf{PE} & All & Known & Novel\\
\midrule
Zero & 89.84 & 84.32 & 92.59\\
Bilinear & 90.84 & 84.86 & 93.83\\
\rowcolor{green!20} Bicubic & \textbf{91.35} & \textbf{86.19} & \textbf{93.93}\\
\bottomrule
\end{tabular}}
\vspace{-2.85em}
\end{wraptable}
\paragraph{Effects of Subtoken PE.} Tab.~\ref{tab:ablation_pe} varies the PEs of the added subtokens in CUB Stage 1, keeping base-token and CLS PEs unchanged. Bilinear and zero PEs lower \textit{All} by $0.51$ and $1.51$ points, respectively. Zero PEs discard where each subtoken lies within its parent patch, indicating that subtokens need position-aware embeddings to relate fine details to the global layout.

\paragraph{Attention Map Comparison.} Fig.~\ref{fig:attnmap_comparison} compares the last-layer CLS attention of pretrained DINOv2, SelEx, and \ours on the three fine-grained datasets. Both baselines allocate a non-negligible portion of attention to background context, \eg, the wire supporting the bird, the ground clutter around the aircraft, and the vegetation behind the car, and their responses remain relatively diffuse across the token grid, as also reflected in the head-averaged maps. In contrast, \ours attends more precisely to the fine-grained patches of the object itself, such as the bird's body, the aircraft fuselage, and the front of the car. Moreover, the attention of \ours is markedly more concentrated on these regions: high responses form compact clusters around discriminative parts rather than spreading over many weakly activated background tokens, indicating a stronger and more consistent focus on class-relevant evidence. We attribute this behavior to subtoken fine-tuning, which repeatedly exposes the model to subdivided discriminative regions and thereby encourages the attention to converge on the local cues that matter for fine-grained recognition.

\paragraph{Fine-grained Details Analysis.} Fig.~\ref{fig:selection_visualization} shows that the selected regions remain semantically coherent across instances despite changes in pose, viewpoint, and background. The emphasized regions also vary with the visual cues required by each domain: bird examples concentrate on wing boundaries and small anatomical parts, aircraft examples emphasize engines and wing structures, and vehicle examples attend to headlights and front-body details. This behavior suggests that the router does not learn a dataset-independent center bias; instead, it adapts the subdivision budget to category-relevant evidence while retaining the base tokens for global context. Fig.~\ref{fig:head_distribution_visualization} further shows that selection frequency is non-uniform across attention heads and differs among datasets, consistent with complementary head semantics. Some heads are rarely selected, which suggests possible redundancy, although establishing whether they can be safely pruned requires a dedicated study.

\section{Conclusion} \label{sec:conclusion}

This work presents \ours, a novel Subtoken Vision Transformer that enhances fine-grained recognition through selective image tokenization. By developing attention-based token subdivision and a two-stage training strategy, our method enables localized resolution enhancement in discriminative regions while maintaining global contextual understanding, bridging pretrained representations and downstream tasks without modifying the backbone architecture. The distilled single-map router further converts attention-guided exploration into a deterministic selection rule, retaining near-baseline inference cost, and its scores further enable inference-time pruning of low-scored base tokens. Experiments on three fine-grained GCD benchmarks show consistent gains over tokenization baselines, especially on novel categories ($+3.4\%$ average \textit{Novel} accuracy over DINOv2), and the learned selection patterns concentrate on semantically coherent object parts, although they are not intended to imply correspondence with human attention.



%
%
\bibliographystyle{splncs04}
\bibliography{main}

@String(CVPR= {IEEE Conf. Comput. Vis. Pattern Recog.})

@String(ICCV= {Int. Conf. Comput. Vis.})

@String(ECCV= {Eur. Conf. Comput. Vis.})

@String(TIP  = {IEEE Trans. Image Process.})

@String(ICLR = {Int. Conf. Learn. Represent.})

@String(AAAI = {AAAI})

@String(CVPRW= {IEEE Conf. Comput. Vis. Pattern Recog. Worksh.})

@String(CVPR  = {CVPR})

@String(ICCV  = {ICCV})

@String(ECCV  = {ECCV})

@String(TIP   = {IEEE TIP})

@String(ICLR  = {ICLR})

@String(CVPRW= {CVPRW})

@article{rao2021dynamicvit,
  title={{DynamicViT}: Efficient vision transformers with dynamic token sparsification},
  author={Rao, Yongming and Zhao, Wenliang and Liu, Benlin and Lu, Jiwen and Zhou, Jie and Hsieh, Cho-Jui},
  journal={NeurIPS},
  year={2021}
}

@inproceedings{havtorn2023msvit,
  title={{MSViT}: Dynamic mixed-scale tokenization for vision transformers},
  author={Havtorn, Jakob Drachmann and Royer, Am{\'e}lie and Blankevoort, Tijmen and Bejnordi, Babak Ehteshami},
  booktitle={ICCVW},
  year={2023}
}

@inproceedings{kim2025sapiensid,
  title={{SapiensID}: Foundation for Human Recognition},
  author={Kim, Minchul and Ye, Dingqiang and Su, Yiyang and Liu, Feng and Liu, Xiaoming},
  booktitle={CVPR},
  year={2025}
}

@inproceedings{wang2021pyramid,
  title={Pyramid vision transformer: A versatile backbone for dense prediction without convolutions},
  author={Wang, Wenhai and Xie, Enze and Li, Xiang and Fan, Deng-Ping and Song, Kaitao and Liang, Ding and Lu, Tong and Luo, Ping and Shao, Ling},
  booktitle={ICCV},
  year={2021}
}

@inproceedings{liu2021swin,
  title={Swin transformer: Hierarchical vision transformer using shifted windows},
  author={Liu, Ze and Lin, Yutong and Cao, Yue and Hu, Han and Wei, Yixuan and Zhang, Zheng and Lin, Stephen and Guo, Baining},
  booktitle={ICCV},
  year={2021}
}

@inproceedings{rusanovsky2024capex,
  title={{CapeX}: Category-Agnostic Pose Estimation from Textual Point Explanation},
  author={Rusanovsky, Matan and Hirschorn, Or and Avidan, Shai},
  booktitle={ICLR},
  year={2025}
}

@inproceedings{chefer2021transformer,
  title={Transformer interpretability beyond attention visualization},
  author={Chefer, Hila and Gur, Shir and Wolf, Lior},
  booktitle={CVPR},
  year={2021}
}

@inproceedings{dosovitskiy2020image,
  title={An image is worth 16x16 words: Transformers for image recognition at scale},
  author={Dosovitskiy, Alexey and Beyer, Lucas and Kolesnikov, Alexander and Weissenborn, Dirk and Zhai, Xiaohua and Unterthiner, Thomas and Dehghani, Mostafa and Minderer, Matthias and Heigold, Georg and Gelly, Sylvain and others},
  booktitle={ICLR},
  year={2021}
}

@inproceedings{zhou2016learning,
  title={Learning deep features for discriminative localization},
  author={Zhou, Bolei and Khosla, Aditya and Lapedriza, Agata and Oliva, Aude and Torralba, Antonio},
  booktitle={CVPR},
  year={2016}
}

@article{jiang2021layercam,
  title={{LayerCAM}: Exploring hierarchical class activation maps for localization},
  author={Jiang, Peng-Tao and Zhang, Chang-Bin and Hou, Qibin and Cheng, Ming-Ming and Wei, Yunchao},
  journal={TIP},
  year={2021}
}

@inproceedings{chowdhury2025prompt,
  title={{Prompt-CAM}: Making Vision Transformers Interpretable for Fine-Grained Analysis},
  author={Chowdhury, Arpita and Paul, Dipanjyoti and Mai, Zheda and Gu, Jianyang and Zhang, Ziheng and Mehrab, Kazi Sajeed and Campolongo, Elizabeth G and Rubenstein, Daniel and Stewart, Charles V and Karpatne, Anuj and others},
  booktitle={CVPR},
  year={2025}
}

@article{han2021autonovel,
  title={{AutoNovel}: Automatically discovering and learning novel visual categories},
  author={Han, Kai and Rebuffi, Sylvestre-Alvise and Ehrhardt, Sebastien and Vedaldi, Andrea and Zisserman, Andrew},
  journal={TPAMI},
  year={2021}
}

@inproceedings{yu2021ap,
  title={{AP-10K}: A benchmark for animal pose estimation in the wild},
  author={Yu, Hang and Xu, Yufei and Zhang, Jing and Zhao, Wei and Guan, Ziyu and Tao, Dacheng},
  booktitle={NeurIPS},
  year={2021}
}

@article{yang2022apt,
  title={{APT-36K}: A large-scale benchmark for animal pose estimation and tracking},
  author={Yang, Yuxiang and Yang, Junjie and Xu, Yufei and Zhang, Jing and Lan, Long and Tao, Dacheng},
  journal={NeurIPS},
  year={2022}
}

@article{fang2022alphapose,
  title={{AlphaPose}: Whole-body regional multi-person pose estimation and tracking in real-time},
  author={Fang, Hao-Shu and Li, Jiefeng and Tang, Hongyang and Xu, Chao and Zhu, Haoyi and Xiu, Yuliang and Li, Yong-Lu and Lu, Cewu},
  journal={TPAMI},
  year={2022}
}

@inproceedings{cao2017realtime,
  title={Realtime multi-person {2D} pose estimation using part affinity fields},
  author={Cao, Zhe and Simon, Tomas and Wei, Shih-En and Sheikh, Yaser},
  booktitle={CVPR},
  year={2017}
}

@inproceedings{zhai2023sigmoid,
  title={Sigmoid loss for language image pre-training},
  author={Zhai, Xiaohua and Mustafa, Basil and Kolesnikov, Alexander and Beyer, Lucas},
  booktitle={ICCV},
  year={2023}
}

@inproceedings{radford2021learning,
  title={Learning transferable visual models from natural language supervision},
  author={Radford, Alec and Kim, Jong Wook and Hallacy, Chris and Ramesh, Aditya and Goh, Gabriel and Agarwal, Sandhini and Sastry, Girish and Askell, Amanda and Mishkin, Pamela and Clark, Jack and others},
  booktitle={ICML},
  year={2021}
}

@inproceedings{kim2024keypoint,
  title={Keypoint relative position encoding for face recognition},
  author={Kim, Minchul and Su, Yiyang and Liu, Feng and Jain, Anil and Liu, Xiaoming},
  booktitle={CVPR},
  year={2024}
}

@inproceedings{teepe2022towards,
  title={Towards a deeper understanding of skeleton-based gait recognition},
  author={Teepe, Torben and Gilg, Johannes and Herzog, Fabian and H{\"o}rmann, Stefan and Rigoll, Gerhard},
  booktitle={CVPRW},
  year={2022}
}

@inproceedings{yang2021transpose,
  title={{TransPose}: Keypoint localization via transformer},
  author={Yang, Sen and Quan, Zhibin and Nie, Mu and Yang, Wankou},
  booktitle={ICCV},
  year={2021}
}

@inproceedings{rastegar2024selex,
  title={{SelEx}: Self-expertise in Fine-Grained Generalized Category Discovery},
  author={Rastegar, Sarah and Salehi, Mohammadreza and Asano, Yuki M and Doughty, Hazel and Snoek, Cees GM},
  booktitle={ECCV},
  year={2024}
}

@inproceedings{vaze2022generalized,
  title={Generalized category discovery},
  author={Vaze, Sagar and Han, Kai and Vedaldi, Andrea and Zisserman, Andrew},
  booktitle={CVPR},
  year={2022}
}

@inproceedings{fini2021unified,
  title={A unified objective for novel class discovery},
  author={Fini, Enrico and Sangineto, Enver and Lathuili{\`e}re, St{\'e}phane and Zhong, Zhun and Nabi, Moin and Ricci, Elisa},
  booktitle={ICCV},
  year={2021}
}

@inproceedings{cao2021open,
  title={Open-world semi-supervised learning},
  author={Cao, Kaidi and Brbic, Maria and Leskovec, Jure},
  booktitle={ICLR},
  year={2022}
}

@inproceedings{deng2009imagenet,
  title={{ImageNet}: A large-scale hierarchical image database},
  author={Deng, Jia and Dong, Wei and Socher, Richard and Li, Li-Jia and Li, Kai and Fei-Fei, Li},
  booktitle={CVPR},
  year={2009}
}

@techreport{wah_branson_welinder_perona_belongie_2011,
  title={The {Caltech-UCSD Birds-200-2011} Dataset},
  author={Wah, Catherine and Branson, Steve and Welinder, Peter and Perona, Pietro and Belongie, Serge},
  institution={California Institute of Technology},
  year={2011}
}

@inproceedings{krause20133d,
  title={{3D} object representations for fine-grained categorization},
  author={Krause, Jonathan and Stark, Michael and Deng, Jia and Fei-Fei, Li},
  booktitle={ICCVW},
  year={2013}
}

@article{maji2013fine,
  title={Fine-grained visual classification of aircraft},
  author={Maji, Subhransu and Rahtu, Esa and Kannala, Juho and Blaschko, Matthew and Vedaldi, Andrea},
  journal={arXiv preprint arXiv:1306.5151},
  year={2013}
}

@inproceedings{caron2021emerging,
  title={Emerging properties in self-supervised vision transformers},
  author={Caron, Mathilde and Touvron, Hugo and Misra, Ishan and J{\'e}gou, Herv{\'e} and Mairal, Julien and Bojanowski, Piotr and Joulin, Armand},
  booktitle={ICCV},
  year={2021}
}

@article{zhao2021novel,
  title={Novel visual category discovery with dual ranking statistics and mutual knowledge distillation},
  author={Zhao, Bingchen and Han, Kai},
  journal={NeurIPS},
  year={2021}
}

@inproceedings{zhong2021openmix,
  title={{OpenMix}: Reviving known knowledge for discovering novel visual categories in an open world},
  author={Zhong, Zhun and Zhu, Linchao and Luo, Zhiming and Li, Shaozi and Yang, Yi and Sebe, Nicu},
  booktitle={CVPR},
  year={2021}
}

@inproceedings{parkhi2012cats,
  title={Cats and dogs},
  author={Parkhi, Omkar M and Vedaldi, Andrea and Zisserman, Andrew and Jawahar, CV},
  booktitle={CVPR},
  year={2012}
}

@article{oquab2024dinov,
  title={{DINO}v2: Learning Robust Visual Features without Supervision},
  author={Maxime Oquab and Timoth{\'e}e Darcet and Th{\'e}o Moutakanni and Huy V. Vo and Marc Szafraniec and Vasil Khalidov and Pierre Fernandez and Daniel Haziza and Francisco Massa and Alaaeldin El-Nouby and Mido Assran and Nicolas Ballas and Wojciech Galuba and Russell Howes and Po-Yao Huang and Shang-Wen Li and Ishan Misra and Michael Rabbat and Vasu Sharma and Gabriel Synnaeve and Hu Xu and Herve Jegou and Julien Mairal and Patrick Labatut and Armand Joulin and Piotr Bojanowski},
  journal={TMLR},
  year={2024}
}

@article{wright1990speeding,
  title={Speeding up the {Hungarian} algorithm},
  author={Wright, Mike B},
  journal={Comput. Oper. Res.},
  year={1990}
}

@inproceedings{zhu2025quality,
  title={A quality-guided mixture of score-fusion experts framework for human recognition},
  author={Zhu, Jie and Su, Yiyang and Kim, Minchul and Jain, Anil and Liu, Xiaoming},
  booktitle={ICCV},
  year={2025}
}

@article{liu2025person,
  title={Person recognition at altitude and range: Fusion of face, body shape and gait},
  author={Liu, Feng and Chimitt, Nicholas and Guo, Lanqing and Jain, Jitesh and Kane, Aditya and Kim, Minchul and Robbins, Wes and Su, Yiyang and Ye, Dingqiang and Zhang, Xingguang and others},
  journal={arXiv preprint arXiv:2505.04616},
  year={2025}
}

@inproceedings{chen2023atm,
  title={{ATM}: Action temporality modeling for video question answering},
  author={Chen, Junwen and Zhu, Jie and Kong, Yu},
  booktitle={ACM MM},
  year={2023}
}

@article{guo2026holistic,
  title={On the Holistic Approach for Detecting Human Image Forgery},
  author={Guo, Xiao and Zhu, Jie and Jain, Anil and Liu, Xiaoming},
  journal={arXiv preprint arXiv:2601.04715},
  year={2026}
}

@inproceedings{zhu2026fusionagent,
  title={{FusionAgent}: A Multimodal Agent with Dynamic Model Selection for Human Recognition},
  author={Zhu, Jie and Guo, Xiao and Su, Yiyang and Jain, Anil and Liu, Xiaoming},
  booktitle={CVPR},
  year={2026}
}

@inproceedings{nilsback2008automated,
  title={Automated flower classification over a large number of classes},
  author={Nilsback, Maria-Elena and Zisserman, Andrew},
  booktitle={ICVGIP},
  year={2008}
}

@article{su2026localscore,
  title={{LocalScore}: Local Density-Aware Similarity Scoring for Biometrics},
  author={Su, Yiyang and Kim, Minchul and Zhu, Jie and Perry, Christopher and Liu, Feng and Jain, Anil and Liu, Xiaoming},
  journal={arXiv preprint arXiv:2602.01012},
  year={2026}
}

@inproceedings{Zhu_2026_CVPR,
  title={Can Textual Reasoning Improve the Performance of {MLLMs} on Fine-Grained Visual Classification?},
  author={Zhu, Jie and Su, Yiyang and Liu, Xiaoming},
  booktitle={CVPR Findings},
  year={2026}
}

@article{zhu2026depthagent,
  title={{DepthAgent}: Towards Better Universal Depth Estimation via Sample-wise Expert Selection},
  author={Zhu, Jie and Ganesan, Girish Chandar and Liu, Xiaoming},
  journal={arXiv preprint arXiv:2605.23281},
  year={2026}
}

@inproceedings{ryoo2021tokenlearner,
  title={{TokenLearner}: Adaptive Space-Time Tokenization for Videos},
  author={Ryoo, Michael S. and Piergiovanni, AJ and Arnab, Anurag and Dehghani, Mostafa and Angelova, Anelia},
  booktitle={NeurIPS},
  year={2021}
}

@inproceedings{yin2022avit,
  title={{A-ViT}: Adaptive Tokens for Efficient Vision Transformer},
  author={Yin, Hongxu and Vahdat, Arash and Alvarez, Jose M. and Mallya, Arun and Kautz, Jan and Molchanov, Pavlo},
  booktitle={CVPR},
  year={2022}
}

@inproceedings{liang2022evit,
  title={Not All Patches are What You Need: Expediting Vision Transformers via Token Reorganizations},
  author={Liang, Youwei and Ge, Chongjian and Tong, Zhan and Song, Yibing and Wang, Jue and Xie, Pengtao},
  booktitle={ICLR},
  year={2022}
}

@article{ma2023dit,
  title={{DiT}: Efficient Vision Transformers with Dynamic Token Routing},
  author={Ma, Yuchen and Fei, Zhengcong and Huang, Junshi},
  journal={arXiv preprint arXiv:2308.03409},
  year={2023}
}

@inproceedings{jiang2024mpfgvc,
  title={Delving into Multimodal Prompting for Fine-Grained Visual Classification},
  author={Jiang, Xin and Tang, Hao and Gao, Junyao and Du, Xiaoyu and He, Shengfeng and Li, Zechao},
  booktitle={AAAI},
  year={2024}
}

@inproceedings{sun2023fgvpl,
  title={Fine-Grained Visual Prompt Learning of Vision-Language Models for Image Recognition},
  author={Sun, Hongbo and He, Xiangteng and Zhou, Jiahuan and Peng, Yuxin},
  booktitle={ACM MM},
  year={2023}
}

@inproceedings{wei2024cascadevlm,
  title={Enhancing Fine-Grained Image Classifications via Cascaded Vision Language Models},
  author={Wei, Canshi},
  booktitle={EMNLP Findings},
  year={2024}
}

@inproceedings{he2025finedefics,
  title={Analyzing and Boosting the Power of Fine-Grained Visual Recognition for Multi-modal Large Language Models},
  author={He, Hulingxiao and Li, Geng and Geng, Zijun and Xu, Jinglin and Peng, Yuxin},
  booktitle={ICLR},
  year={2025}
}

@inproceedings{xie2025fgclip,
  title={{FG-CLIP}: Fine-Grained Visual and Textual Alignment},
  author={Xie, Chunyu and Wang, Bin and Kong, Fanjing and Li, Jincheng and Liang, Dawei and Zhang, Gengshen and Leng, Dawei and Yin, Yuhui},
  booktitle={ICML},
  year={2025}
}

@inproceedings{xiao2025flair,
  title={{FLAIR}: {VLM} with Fine-Grained Language-Informed Image Representations},
  author={Xiao, Rui and Kim, Sanghwan and Georgescu, Mariana-Iuliana and Akata, Zeynep and Alaniz, Stephan},
  booktitle={CVPR},
  year={2025}
}

@inproceedings{weber2000unsupervised,
  title={Unsupervised learning of models for recognition},
  author={Weber, Markus and Welling, Max and Perona, Pietro},
  booktitle={ECCV},
  year={2000}
}

@inproceedings{fergus2003object,
  title={Object class recognition by unsupervised scale-invariant learning},
  author={Fergus, Rob and Perona, Pietro and Zisserman, Andrew},
  booktitle={CVPR},
  year={2003}
}

@article{felzenszwalb2010object,
  title={Object detection with discriminatively trained part-based models},
  author={Felzenszwalb, Pedro F. and Girshick, Ross B. and McAllester, David and Ramanan, Deva},
  journal={TPAMI},
  year={2010}
}

@inproceedings{zhang2013deformable,
  title={Deformable part descriptors for fine-grained recognition and attribute prediction},
  author={Zhang, Ning and Farrell, Ryan and Iandola, Forrest and Darrell, Trevor},
  booktitle={ICCV},
  year={2013}
}

@inproceedings{bourdev2009poselets,
  title={Poselets: Body part detectors trained using {3D} human pose annotations},
  author={Bourdev, Lubomir and Malik, Jitendra},
  booktitle={ICCV},
  year={2009}
}

@inproceedings{zhang2014part,
  title={Part-based {R-CNNs} for fine-grained category detection},
  author={Zhang, Ning and Donahue, Jeff and Girshick, Ross and Darrell, Trevor},
  booktitle={ECCV},
  year={2014}
}

@inproceedings{su2026sapiensid2,
  title={{SapiensID} 2.0: Aligning Human Recognition Foundation Models with Human Perception},
  author={Su, Yiyang and Zhu, Jie and Liu, Feng and Jain, Anil K. and Liu, Xiaoming},
  booktitle={NeurIPS},
  year={2026}
}

@inproceedings{xu2026emotag,
  title={{EmoTaG}: Emotion-Aware Talking Head Synthesis on {Gaussian} Splatting with Few-Shot Personalization},
  author={Xu, Haolan and Cheng, Keli and Wang, Lei and Bi, Ning and Liu, Xiaoming},
  booktitle={CVPR},
  year={2026}
}

@inproceedings{ding2019selective,
  title={Selective sparse sampling for fine-grained image recognition},
  author={Ding, Yao and Zhou, Yanzhao and Zhu, Yi and Ye, Qixiang and Jiao, Jianbin},
  booktitle={ICCV},
  year={2019}
}

@inproceedings{ren2023chatgpt,
  author = {Zhiyuan Ren and Yiyang Su and Xiaoming Liu},
  title = {ChatGPT-Powered Hierarchical Comparisons for Image Classification},
  booktitle = {NeurIPS},
  year = {2023},
}

@inproceedings{su2024open,
  author = {Yiyang Su and Minchul Kim and Feng Liu and Anil Jain and Xiaoming Liu},
  title = {Open-Set Biometrics: Beyond Good Closed-Set Models},
  booktitle = {ECCV},
  year = {2024},
}

@inproceedings{liu2024farsight,
  author = {Feng Liu and Ryan Ashbaugh and Nicholas Chimitt and Najmul Hassan and Ali Hassani and Ajay Jaiswal and Minchul Kim and Zhiyuan Mao and Christopher Perry and Zhiyuan Ren and Yiyang Su and Pegah Varghaei and Kai Wang and Xingguang Zhang and Stanley Chan and Arun Ross and Humphrey Shi and Zhangyang Wang and Anil Jain and Xiaoming Liu},
  title = {FarSight: A Physics-Driven Whole-Body Biometric System at Large Distance and Altitude},
  booktitle = {WACV},
  year = {2024},
}

@inproceedings{su2025hamobe,
  author = {Yiyang Su and Yunping Shi and Feng Liu and Xiaoming Liu},
  title = {HAMoBE: Hierarchical and Adaptive Mixture of Biometric Experts for Video-based Person ReID},
  booktitle = {{ICCV}},
  year = {2025},
}


\end{document}